# Improved Adaptive Brovey as a New Method for Image Fusion

Hamid Reza Shahdoosti

*Abstract*— An ideal fusion method preserves the Spectral information in fused image and adds spatial information to it with no spectral distortion. Among the existing fusion algorithms, the contourlet-based fusion method is the most frequently discussed one in recent publications, because the contourlet has the ability to capture and link the point of discontinuities to form a linear structure. The Brovey is a popular pan-sharpening method owing to its efficiency and high spatial resolution. This method can be explained by mathematical model of optical remote sensing sensors. This study presents a new fusion approach that integrates the advantages of both the Brovey and the cotourlet techniques to reduce the color distortion of fusion results. Visual and statistical analyzes show that the proposed algorithm clearly improves the merging quality in terms of: correlation coefficient, ERGAS, UIQI, and Q4; compared to fusion methods including IHS, PCA, Adaptive IHS, and Improved Adaptive PCA.

## I. INTRODUCTION

In remote sensing systems, scenes are observed in different portions of electromagnetic spectrum; therefore, the remote sensing images vary in spectral and spatial resolution. To collect more photons and maintain signal to noise ratio simultaneously, the multispectral sensors, with high spectral resolution, have a lower spatial resolution compared with panchromatic image with a higher spatial resolution, and wide spectral bandwidth. With proper algorithms, it is possible to fuse these images and produce imagery with the best information of both, namely high spatial and high spectral resolution.

The most commonly used image-fusion methods are those based on the IHS (also known as HSI) [1], and principal component analysis (PCA) [2], [3]. However, these methods can cause spectral distortion in the results [2]. Chavez proposed HPF(High Pass Filtering) fusion algorithm which has shown better performance in terms of high-quality synthesis of spectral information [4]. Principle of this method is to extract the high-frequency information from the PAN image and inject it into the low resolution multispectral (LMS) image, formerly resampled to match the PAN pixel size. Box car filters are used as a low pass filter in this method. However, the ripple in the frequency response of box car filters has some negative impact on this method.

The well-known Mallat's fusion algorithm [5] uses an orthonormal basis and can effectively preserve spectral information , but because of down-sampling operators, which are used to implement ordinary discrete wavelet transforms, these transforms are not shift-invariant and can lead to problem in data fusion [6].

To avoid above-mentioned problem, the discrete wavelet transform known as "à trous" algorithm [7] was proposed by eliminating the decimation operators in wavelet structure. It is a shift-invariant and redundant wavelet transform algorithm based on a multiresolution dyadic scheme. However, wavelets cannot capture the smoothness along the contours [8].

An efficient decomposition of the image should ideally possess properties of multiresolution, localization, directionality, and anisotropy. An alternative multiresolution approach, called contourlet transform, provides an efficient directional decomposition and is also has ability of capturing spatial structures of the natural image along the smooth contours [9].

The Brovey, based on the chromaticity transform [10], [11], [12], is a simple method for combining data from different sensors. It also preserves a high degree of spatial detail of the PAN image; however, distortion of the spectral information is not acceptable at all.

To overcome the deficiency of Brovey method, a new fusion approach is proposed in this letter. We call this algorithm "Adaptive-Brovey" because it varies the amount of spatial injection depending on the initial LMS and panchromatic (PAN) images. To minimize spectral distortion, we use NSCT (Nonsubsampled Contourlet Transform) to add higher frequency components of Adaptive Brovey result to the initial LMS images. As we know, all of the lower frequency components of HMS (High Resolution MS) are available in the LMS image and we can use this spectral information to improve Adaptive Brovey.

This improved adaptive Brovey method integrates the advantages of both the Brovey and the NSCT techniques to preserve both spatial and spectral information of PAN and LMS images. To verify the efficiency of the proposed method, visual and statistical assessments are carried out on LMS and PAN data.

## II. ADAPTIVE BROVEY ALGORITHM

Image formation model shows that the ratio between $DN_b^h$ and $DN_b^L$ is independent of the sensor band [13]. If PAN represents a panchromatic band and LMS a Low resolution multispectral band, then we will have:

$$DN_{MS_i}^h = \frac{DN_{PAN}^h}{DN_{PAN}^L} DN_{MS_i}^L \quad (1)$$

where $DN_{PAN}^h$ and $DN_{MS}^L$ are pixel values of the PAN and the corresponding pixel values of the LRM, respectively. Different methods are used to compute $DN_{PAN}^L$ in [13-15]. The $DN_{PAN}^L$ is estimated from $DN_{MS}^L$ in the Brovey method. So we will have:

$$DN_{MS_i}^h = \frac{DN_{PAN}^h}{\frac{1}{N}\sum_{i=1}^{N} DN_{MS_i}^L} DN_{MS_i}^L \quad (2)$$

where $DN_{MS_i}^h$ is the i$^{th}$ fused multispectral band at the resolution of Pan image. When Brovey is used, injection of spatial information into the LMS images is more than desirable, leading to the spectral distortion in the results [13].

The ratio between $DN_{PAN}^h$ and $\frac{1}{N}\sum_{i=1}^{N} DN_{MS_i}^l$ in equation (2) is the gain of sharpening. To avoid over-sharpening of the Brovey method, we propose the Adaptive Brovey to control the spatial information of the fused image

$$DN_{MS_i}^h = \left(\frac{DN_{PAN}^h}{\sum_{i=1}^{N} b_i \times DN_{MS_i}^L}\right)^a \times DN_{MS_i}^L \quad (3)$$

where $b_i$ values are determined by minimizing the RMSE between $\sum_{i=1}^{N} b_i \times DN_{MS_i}^L$ and PAN image, using method introduced in [16], and $a$ is the parameter which can control the injection of spatial information. If $a$ equals zero, no fusion is performed and if $a$ equals one, equation (2) and (3) are the same.

To find a suitable $a$ parameter, we vary the value of $a$ from zero to one by an arbitrary step like 0.05 and fuse the PAN and LMS images by equation (3). The best $a$ value is determined by minimizing the QNR index. Recently proposed by Alparone *et al.* [17], the QNR index evaluates the quality of the fused image without requiring the HRM image and combines the two distortion indexes of the radiometric and geometric distortion indexes. In this manner, fused image with simultaneously high spectral and spatial resolution can be achieved. The main advatntage of Brovey on which we emphasize is that the spatial information of each fused pixel is relative to its value. This means that Brovey uses local gain in its fusion procedure. For example, vegetation has almost no response in the blue range, which exhibits dark digits in (multispectral) MS images [18]. The fused result of equation (4) or (5) remains zero in this case (vegetation area in the blue range), but other methods inject spatial information of other bands into the blue band in this example. The context-based-driven algorithms (CBD) [19] try to add spatial information relative to its local energy, whereas the applied window which is used in such algorithms takes into account neighbor pixels and may lead to spectral and spatial distortion in some cases such as vegetation area and around it.

## III. NSCT Transform

This section reviews the NSCT as presented in [9] and [20]. The nonsubsampled contourlet transform (NSCT) is used to decompose input images into low-frequency and high-frequency parts. The contourlet has the ability to capture and link the point of discontinuities to form a linear structure or contours. NSCT can be divided into two shift-invariant parts: 1) nonsubsampled pyramid structure (NSP) and 2) a nonsubsampled directional filter bank structure (NSDFB).

1) NSP: NSP is achieved by using two-channel nonsubsampled 2-D filter banks. This expansion is conceptually similar to the 1-D NSWT computed with the ATWT (à trous wavelet Transform) algorithm and has n+1 redundancy, where n is the number of decomposition stages. Filters of the subsequent stages are upsampled version of those of the first stage. The desired passband support of the low-pass filter at the nth stage is the interval $[-(\frac{p}{2^n}),(\frac{p}{2^n})]^2$. Accordingly, the ideal region of the equivalent high-pass filter is the complement of the low-pass one, i.e., the interval $[-(\frac{p}{2^{n-1}}),(\frac{p}{2^{n-1}})]^2 / [-(\frac{p}{2^n}),(\frac{p}{2^n})]^2$. Without using any additional filter design, this gives a multiscale property. At each stage one bandpass image is produced. Thus, certain parts of the noise spectrum are filtered in the processed pyramid coefficients.

2) NSDFB: The NSDFB is constructed by eliminating the downsamplers and upsamplers operator in the directional filter bank (DFB). By switching off the downsamplers/upsamplers in each two-channel filter bank in the DFB tree structure and upsampling the filters accordingly, a shift-invariant directional expansion is obtained. These results in a tree are composed of two-channel NSFBs. By combining the NSP and the NSDFB, the NSCT is constructed.

## IV. Improved Adaptive Brovey

In the past few years, several researchers proposed different PAN and MS image-fusion methods based on hybrid concept to integrate the advantages of two fusion methods. Undecimated Wavelet PCA (udWPC) [12], Undecimated Wavelet IHS (udWI) [12], IHS-wavelet [2], IHS- retina-inspired [21], FFT-enhanced IHS [22], and Improved adaptive PCA [8] are the most commonly used hybrid methods. However, Adaptive Brovey tries to preserve spectral and spatial information simultaneously and it is faster than other similar algorithms such as [20].

In order to minimize spectral distortion in the Adaptive Brovey method, we use NSCT to preserve the lower frequency of LMS images. As we know, the missed high frequency components of HMS images are only needed to improve the spatial resolution of LMS images, so we use NSCT to decompose LMS images and fused result of Adaptive Brovey into low-frequency and high-frequency parts. Improved Adaptive Brovey replaces the detail contourlet coefficients of the LMS result by those of Adaptive Brovey images. Because higher frequency components of images have zero mean, using the lower frequency component of LMS (which is equal in LMS and reference image) in any fusion method can at least remove bias between Fused result and the reference image. One can find a suitable $a$ parameter by varying the value of $a$ and fuse the PAN and LMS images by Improved Adaptive Brovey. The best $a$ value is determined by minimizing the QNR index between Improved Adaptive Brovey result and initial LMS and PAN images.

## V. EXPERIMENTAL RESULTS

Four-band multispectral QuickBird data was used for our experiments. These images are acquired by a commercial satellite, QuickBird. The QuickBird data set was taken over the Pyramid area of Egypt in 2002. The subscenes of the raw images are used as PAN and LMS.

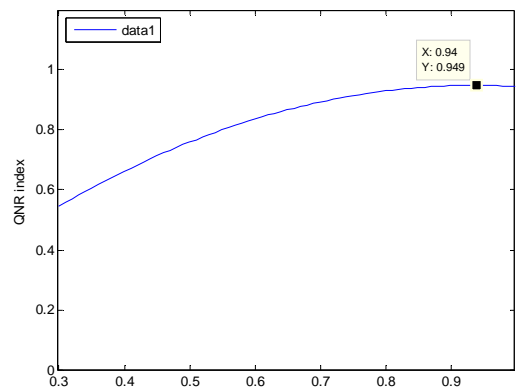

Fig.1. QNR index Curve versus $a$

Two recent adaptive methods considered for comparison are Adaptive IHS and Improved Adaptive PCA. To implement Adaptive IHS method, we use software which is introduced in [16]. A MATLAB toolbox that implements the NSCT is used for implementation of Improved Adaptive PCA and Improved Adaptive Brovey. Original PAN and LMS images are spatially degraded down to a lower resolution in order to compare fused product to the genuine references resolution.

Fig. 1 shows the QNR index versus $a$ in our experiment. As can be seen from Fig. 1, the value of $a$ which maximize the QNR equals .94.

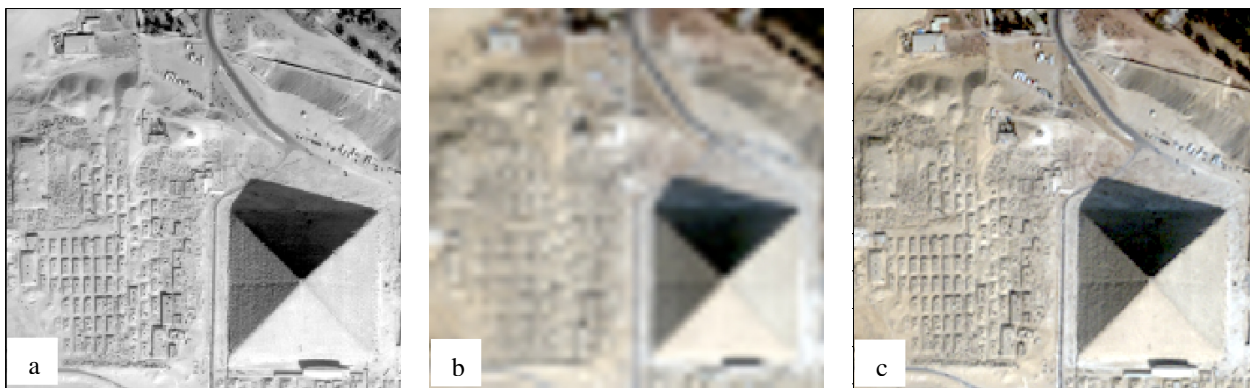

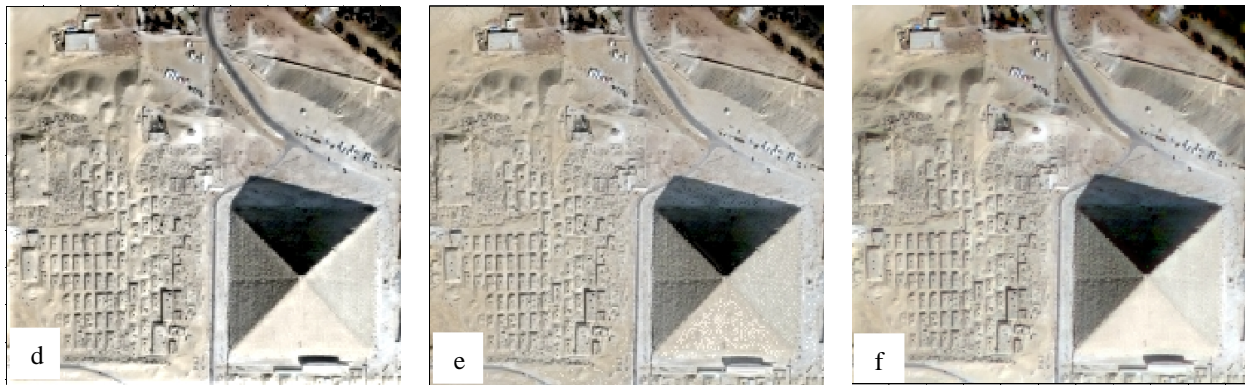

Fig.2. (a) PAN. (b) LMS. (c) Reference MS image. (d) Fused image by Improved Adaptive PCA. (e) Fused image by Adaptive IHS. (f) Fused image by Improved Adaptive Brovey.

TABLE I : Spectral Quality Metrics for Fig. 2

|  | PCA | Improved Adaptive PCA | IHS | Adaptive HIS | Improved Adaptive Brovey |
|---|---|---|---|---|---|
| CC | 0.9125 | 0.9757 | 0.9261 | 0.9517 | 0.9778 |
| ERGAS | 3.9711 | 1.9737 | 3.6568 | 2.4235 | 1.8774 |
| UIQI | 0.8444 | 0.9305 | 0.8439 | 0.9265 | 0.9427 |
| Q4 | 0.8033 | 0.9016 | 0.8128 | 0.8944 | 0.9282 |

Visual comparison of the fused images is the first step of quality assessment. To do this, several aspects of the image quality such as surfaces, linear features, edges, color, blurring, and blooming can be taken into account. The visual performances of the fused images are shown in Fig. 2.

Fig. 2 illustrates fusion results at inferior level. According to Fig. 2, the spatial details pretty well enhanced in the merged image using Improved Adaptive PCA method, but the colors have been changed during the fusion process which presents the spectral distortion. As illustrated in Fig 2, the sharpening in the Adaptive IHS and Improved Adaptive Brovey method is suitable; however, some false edges can be seen in Adaptive IHS at pyramid area. These artifacts probably come from unsuitable parameters of the edge detector (see [16]).

Over-enhanced edges are evident in vegetation area of both Adaptive IHS and Improved Adaptive PCA (upper right corner of the image).

To show similarity between the reference image and fused images, histogram of the reference image and fused images for green band are represented in Fig. 3. The histogram of proposed method is more similar to that of reference image in this figure.

In addition to the visual inspection, the performance of each method should be analyzed quantitatively. In order to assess the quality of the merged images at the inferior level, four objective indicators were used [23-29].

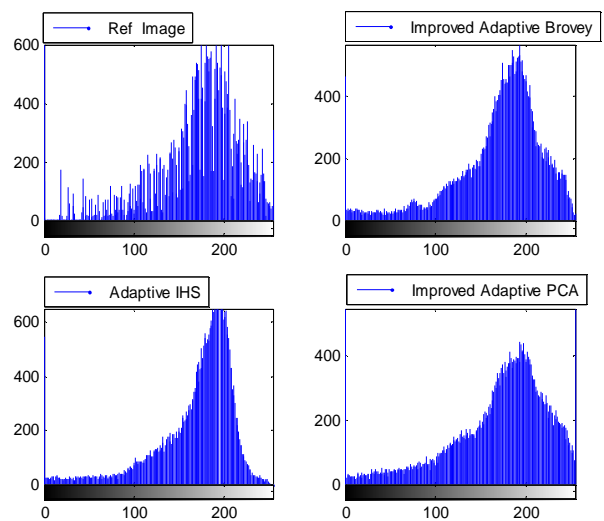

Fig. 3 Histogram of the green band of Fig. 2.

1) Correlation coefficient

This metric indicates the degree of linear dependence between the original reference and fused image. If two images are identical, the correlation coefficient will be maximal and equals 1. It is defined as follows:

$$Corr(A,B) = \frac{\sum_{j=1}^{npix}(A_j - m_A)(B_j - m_B)}{\sqrt{\sum_{j=1}^{npix}(A_j - m_A)\sum_{j=1}^{npix}(B_j - m_B)}}. \quad (4)$$

2) ERGAS

ERGAS (or relative global dimensional synthesis error) is as follows:

$$ERGAS = 100 \frac{h}{l} \sqrt{\frac{1}{N}\sum_{i=1}^{N}\frac{RMSE^2(B_i)}{M_i^2}} \quad (5)$$

where h is the resolution of the high spatial resolution image, l is the resolution of the low spatial resolution image, and $M_i$ is the mean radiance of a specific band involved in the fusion. RMSE is the root mean square error and can be computed by the following expression:

$$RMSE^2(B_i) = Bias^2(B_i) + SD^2(B_i) \quad (6)$$

The lower the value of the ERGAS is, the higher the quality of the merged images is.

3) UIQI

It is defined as follows:

$$Q = \frac{s_{AB}}{s_A s_B} \cdot \frac{2m_A m_B}{m_A^2 + m_B^2} \cdot \frac{2s_A s_B}{s_A^2 + s_B^2} \quad (7)$$

where A and B are fused and reference images respectively. The universal image quality index (UIQI) models any distortion as a combination of three different factors: loss of linear correlation, contrast distortion and luminance distortion.

4) $Q_4$

$Q_4$ is composed of three different factors: The first is the modulus of the hyper-complex CC between the two spectral pixel vectors which is sensitive to both the loss of correlation and to spectral distortion between the two MS data sets. Contrast changes and mean bias on all bands are measured by the second and third factors, respectively.

The Table I shows the values of CC, ERGAS, UIQI and $Q_4$ for proposed method and other mentioned methods. As can be seen from Table I, the quality indexes obtained by applying proposed method are pretty good compared with those obtained by applying other methods. These statistical assessment findings agree with those of the visual analysis. Visual and statistical assessments show that the proposed methods have the least color distortions and desirable spatial information.

I. CONCLUSION

A new pan-sharpening method based on hybrid concept is proposed in this paper. To avoid the over-enhancement of the Brovey method, we proposed the Adaptive Brovey to control the spatial information of the fused image. In order to minimize spectral distortion in the Adaptive Brovey method, the Improved Adaptive Brovey uses NSCT to preserve the lower frequency of LMS images. Proposed Improved Adaptive Brovey method maintains spectral and spatial information simultaneously. Finally, the visual results show that Improved Adaptive Brovey can achieve better performance in comparison with some new methods. In addition to the visual inspection, the performance of proposed method and some former well-known methods were analyzed quantitatively. We applied correlation coefficient, ERGAS, UIQI and Q4 metrics to obtain the quality values. The performance evaluation metrics confirmed the superiority of the Improved Adaptive Brovey method.